\documentclass[11pt]{article}
\usepackage[a4paper, margin=1in]{geometry}
\usepackage[utf8]{inputenc}
\usepackage[T1]{fontenc}
\usepackage{times}
\usepackage[hidelinks]{hyperref}
\usepackage{tabularx}
\usepackage{url}
\usepackage{booktabs}
\usepackage{amsfonts}
\usepackage{amsmath}
\usepackage{amssymb}
\usepackage{nicefrac}
\usepackage{microtype}
\usepackage{xcolor}
\usepackage{graphicx}
\usepackage{enumitem}
\providecommand{\citep}[1]{\cite{#1}}
\providecommand{\citet}[1]{\cite{#1}}
\title{FreqFormer: Hierarchical Frequency-Domain Attention with Adaptive Spectral Routing for Long-Sequence Video Diffusion Transformers  An Algorithm-Architecture Co-Design}
\author{Haopeng Jin \\ Beijing University of Posts and Telecommunications \\ \texttt{Sunbeam.King@bupt.edu.cn}}
\date{Technical Report \\\today}
\begin{document}
\maketitle
\begin{abstract}
Long-sequence video diffusion transformers are constrained by the quadratic cost of self-attention, which becomes severe when sequence lengths reach hundreds of thousands of tokens in long-form video generation. Existing efficient attention methods typically impose a uniform approximation across all content, although video representations exhibit strong spectral structure: low spatial-temporal frequencies often encode global scene layout and coarse motion, while higher frequencies emphasize local texture and fine detail [Torralba and Oliva, 2003; Wang et al., 2023a]. We present \textbf{FreqFormer}, a frequency-aware heterogeneous attention framework for video diffusion transformers. FreqFormer decomposes projected token features into spectral bands and assigns each band a different interaction pattern: dense global attention on compressed low-frequency content, structured block-sparse attention on mid-frequency content, and local sliding-window attention on high-frequency content. A lightweight \textbf{spectral routing network} allocates attention heads across bands as a function of layer statistics and diffusion timestep, enabling compute to shift from global structure modeling in early denoising to detail refinement in later steps, consistent with prior observations on coarse-to-fine denoising dynamics in diffusion models [Ho et al., 2020; Karras et al., 2022]. To preserve coherence across bands, the framework includes low-cost cross-band residual exchange through summary tokens.

FreqFormer is developed as an \textbf{algorithm-architecture co-design}. The algorithm is paired with a fused GPU execution strategy that co-schedules dense, sparse, and local branches to reduce launch overhead and memory traffic, following the broader lesson that attention efficiency depends jointly on algorithmic structure and IO-aware implementation [Dao et al., 2022; Dao, 2024]. We present a consistent complexity framework, an approximation perspective based on orthonormal spectral decompositions, and a simulation-backed systems analysis covering throughput, arithmetic intensity, memory traffic, and duration scaling. Across simulated sequence lengths from 64K to 1M tokens, FreqFormer reduces estimated attention FLOPs by 9.327.3 relative to dense attention and cuts estimated KV-score memory traffic by 8.820.9, while maintaining a hardware-compatible execution pattern. These results support the central claim of the paper: \textbf{spectrally structured heterogeneous attention is a promising and practical direction for long-video diffusion transformers}.
\end{abstract}

\tableofcontents
\newpage
\section{Introduction}
\subsection{Motivation}

Transformers have become a central backbone for diffusion-based generative modeling, including image and video synthesis [Peebles and Xie, 2023; Blattmann et al., 2023; Brooks et al., 2024]. In these systems, self-attention provides flexible long-range interaction but incurs \(O(N^2)\) pairwise cost for a sequence of length \(N\). For video generation, where token counts scale with temporal length and spatial resolution, this cost quickly dominates both runtime and memory.

A broad literature addresses this bottleneck through exact IO-aware kernels [Dao et al., 2022; Dao, 2024], sparse attention [Child et al., 2019; Beltagy et al., 2020; Zaheer et al., 2020], linearized attention [Katharopoulos et al., 2020; Choromanski et al., 2021; Peng et al., 2021], and alternative long-range sequence operators [Lee-Thorp et al., 2022; Poli et al., 2023]. However, most methods impose one interaction rule uniformly over all content. For video diffusion, that uniformity is likely suboptimal because generated video latents and internal features are not spectrally homogeneous.

\subsection{Frequency-dependent structure in video}

Natural videos contain strong low-frequency structure associated with scene layout, object configuration, illumination, and coarse motion, while high-frequency content more often reflects edges, texture, and fine local variation [Torralba and Oliva, 2003; Simoncelli and Olshausen, 2001]. Recent analyses of learned video representations also report nonuniform frequency concentration and scale-dependent temporal dynamics [Feichtenhofer et al., 2022; Tong et al., 2022; Wang et al., 2023a]. This motivates a nonuniform attention policy:

\begin{itemize}[leftmargin=*]
\item \textbf{Low-frequency content} should preserve broad global interaction.
\item \textbf{Mid-frequency content} can often be modeled with structured sparse interaction.
\item \textbf{High-frequency content} is frequently localized and can be handled with local neighborhoods.
\end{itemize}

FreqFormer operationalizes this idea by moving attention into a frequency-aware processing pipeline.

\subsection{Proposed method}

FreqFormer applies a separable spectral transform to per-head query, key, and value features, partitions transformed coefficients into low-, mid-, and high-frequency bands, and applies a different attention operator to each band:

\begin{itemize}[leftmargin=*]
\item \textbf{Dense global attention} on compressed low-band tokens,
\item \textbf{Strided block-sparse attention} on mid-band tokens,
\item \textbf{Sliding-window local attention} on high-band tokens.
\end{itemize}

A \textbf{spectral routing network} adaptively allocates head budget across bands as a function of pooled spectral statistics and diffusion timestep embedding. Band outputs communicate through summary-token residual exchange and are reconstructed through an inverse transform before the output projection.

\subsection{Contributions}

This paper makes the following contributions:

\begin{enumerate}[leftmargin=*]
\small
\item \textbf{Frequency-aware heterogeneous attention.} We propose a transformer attention mechanism that aligns attention structure with spectral bands rather than imposing a single approximation globally.
\item \textbf{Adaptive spectral routing.} We introduce timestep-conditioned head allocation across frequency bands to match the changing requirements of diffusion denoising.
\item \textbf{Algorithm-architecture co-design.} We describe a fused execution strategy for dense, sparse, and local branches inspired by IO-aware attention kernels and hardware-conscious scheduling [Dao et al., 2022; Ham et al., 2024].
\item \textbf{Consistent complexity and approximation analysis.} We provide a parameterized cost model and an output-space approximation perspective under orthonormal transforms.
\item \textbf{Simulation-backed systems evaluation.} We report exact computed values from our simulation setup for FLOPs, memory traffic, arithmetic intensity, throughput estimates, and duration scaling.
\end{enumerate}

\begin{figure}[htbp]
\centering
\includegraphics[width=0.85\linewidth]{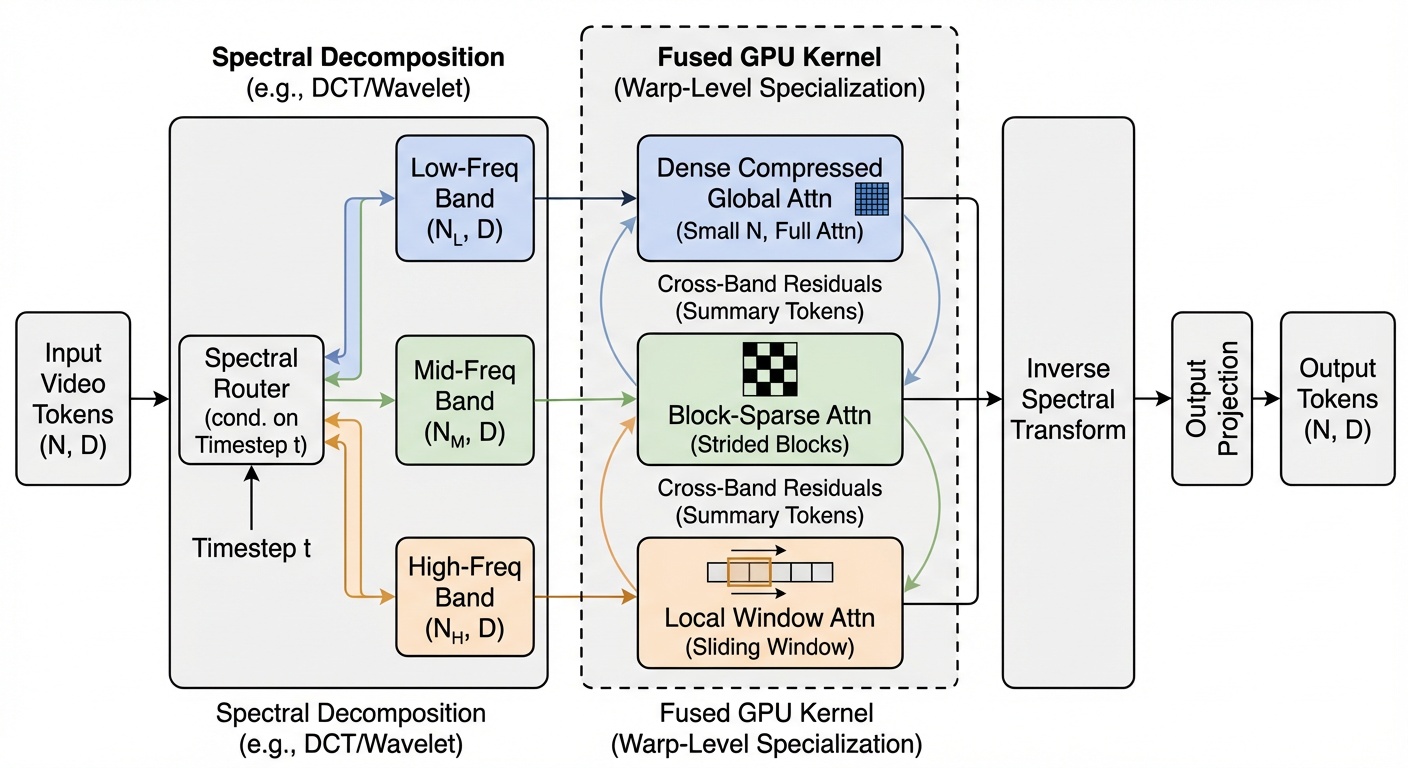}
\caption{\textbf{FreqFormer overview: frequency decomposition, adaptive routing, and heterogeneous attention}. High-level diagram of the FreqFormer layer. Input video tokens are transformed into a spectral basis, partitioned into low-, mid-, and high-frequency bands, and routed to dense compressed global attention, block-sparse attention, and local window attention respectively. A timestep-conditioned router allocates heads across bands, cross-band residual exchange passes summary tokens between bands, and a fused GPU kernel computes all branches before inverse transformation and output projection.}
\label{fig:fig1}
\end{figure}
\section{Related Work}
\subsection{Diffusion transformers for image and video generation}

Diffusion Transformers (DiT) demonstrated that transformer backbones can scale diffusion modeling effectively in image generation [Peebles and Xie, 2023]. This direction has since been extended to video generation and video diffusion systems with increasingly long contexts and larger latent spaces [Blattmann et al., 2023; Ho et al., 2022; Singer et al., 2023; Brooks et al., 2024]. As sequence length grows, full dense attention becomes a key computational bottleneck.

\subsection{Efficient attention and long-context sequence modeling}

Several families of methods reduce the cost of long-context sequence interaction.

\textbf{Exact but IO-aware attention.} FlashAttention and its successors reduce memory traffic and improve throughput without changing the exact attention result [Dao et al., 2022; Dao, 2024].

\textbf{Sparse attention.} Sparse Transformer, Longformer, BigBird, and related methods replace full pairwise connectivity with structured patterns [Child et al., 2019; Beltagy et al., 2020; Zaheer et al., 2020].

\textbf{Linear attention.} Linear Transformers, Performer, and related approaches kernelize or reparameterize attention to avoid quadratic score matrices [Katharopoulos et al., 2020; Choromanski et al., 2021; Peng et al., 2021].

\textbf{Attention alternatives.} Hyena and long convolution methods replace pairwise attention with implicit filters or long convolutions [Poli et al., 2023; Gu et al., 2022].

FreqFormer differs from these methods by using \textbf{different attention operators for different spectral bands}, rather than one approximation for all tokens.

\subsection{Spectral token mixing and spectral operators}

FNet showed that Fourier transforms can act as efficient token mixers in transformer architectures [Lee-Thorp et al., 2022]. Fourier Neural Operator demonstrated that spectral-domain global interactions can be effective in structured domains [Li et al., 2021]. Related work has studied wavelet and frequency-based neural architectures for efficient representation learning [Bruna and Mallat, 2013; Tolstikhin et al., 2021]. FreqFormer builds on this line of reasoning but retains content-dependent Q/K/V attention and uses the spectral domain to determine \textbf{which interaction pattern to apply}.

\subsection{Hybrid attention for video and long-context generation}

Hybrid architectures that combine local and global branches, sparse and dense operators, or multi-scale pathways have been successful in vision and video modeling [Liu et al., 2021; Arnab et al., 2021; Bertasius et al., 2021; Fan et al., 2021]. Recent video diffusion systems also combine local temporal processing, windowing, and sparse long-range interactions to manage cost [Blattmann et al., 2023; Brooks et al., 2024]. FreqFormer is related in spirit but organizes heterogeneity around \textbf{frequency bands} rather than only geometry or scale.

\subsection{Algorithm-kernel co-design}

The systems perspective follows work showing that algorithmic efficiency is insufficient without implementation-level optimization. FlashAttention established the importance of IO-aware tiling and kernel fusion [Dao et al., 2022]. Recent hardware-oriented studies also emphasize pipelined asymmetric kernels, collective-aware scheduling, and many-PE dataflow optimization for transformer workloads [Ham et al., 2024; Kwon et al., 2023; Narayanan et al., 2021]. FreqFormer extends this viewpoint to a heterogeneous multi-pattern attention operator.

\begin{figure}[htbp]
\centering
\includegraphics[width=0.85\linewidth]{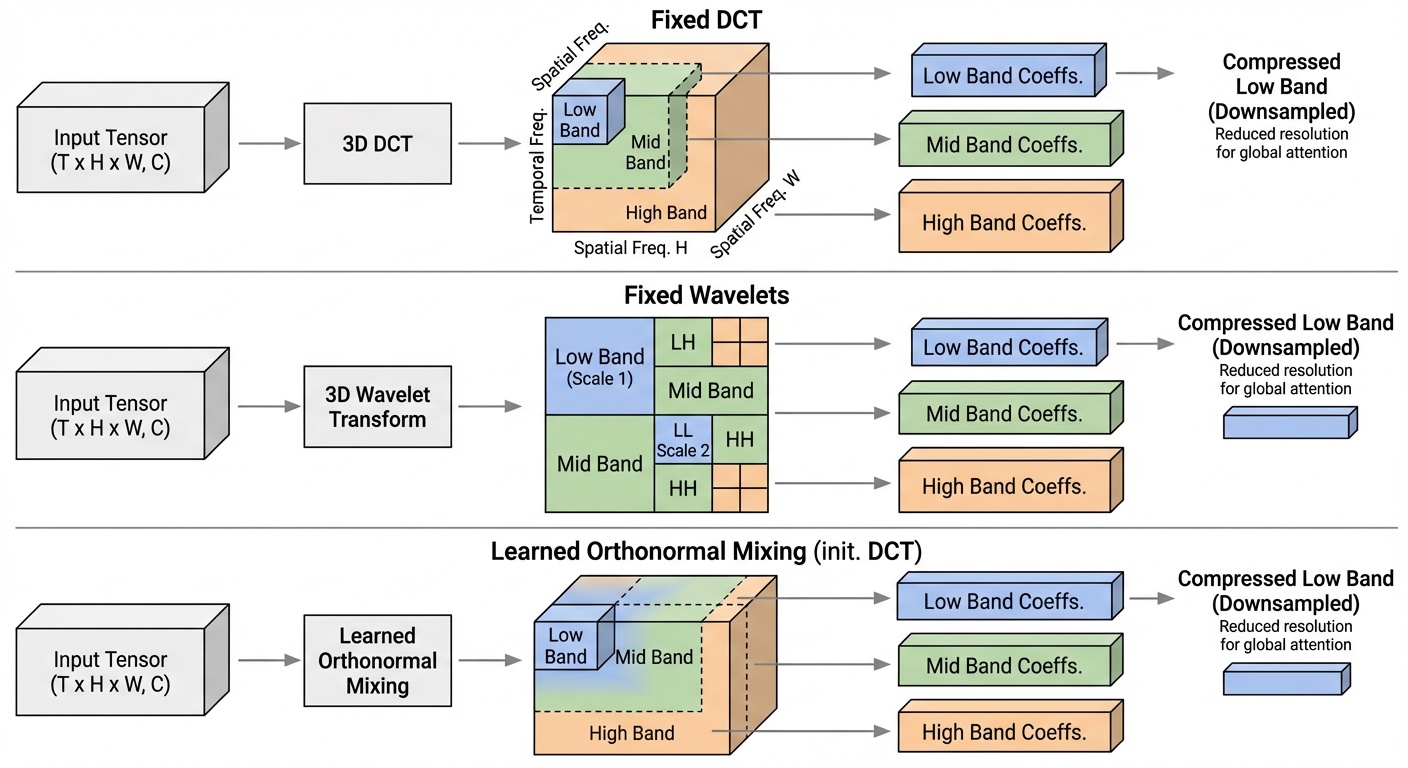}
\caption{\textbf{Learnable spectral decomposition and band partitioning}. Illustration of the spectral decomposition layer with alternative transform instantiations: fixed DCT, fixed wavelets, and learned orthonormal mixing initialized from DCT. The figure shows how transformed coefficients are partitioned into low, mid, and high bands over spatial-temporal token axes, and how low-frequency coefficients are further compressed before attention.}
\label{fig:fig2}
\end{figure}
\section{Method}
\subsection{Setup and notation}

Let the input video latent tensor be
\[
X \in \mathbb{R}^{T_v \times H \times W \times d},
\]
where \(T_v\) is the temporal length, \(H\) and \(W\) are spatial dimensions, and \(d\) is the channel width. The flattened sequence length is
\[
N = T_v H W.
\]

For head \(h \in \{1,\dots,n_h\}\), linear projections produce
\[
Q_h, K_h, V_h \in \mathbb{R}^{T_v \times H \times W \times d_k}.
\]

To avoid notation clashes, we use \(\mathcal{F}\) for the spectral transform operator and reserve \(T_v\) for the temporal dimension.

\subsection{Separable spectral decomposition}

A full learned \(N \times N\) transform is impractical at long sequence lengths. FreqFormer uses a separable transform
\[
\mathcal{F} = F_t \otimes F_h \otimes F_w,
\]
where \(F_t\), \(F_h\), and \(F_w\) are 1D transforms along temporal, height, and width axes, respectively. In the reference design these are fixed orthonormal DCT-II transforms, though wavelets are also possible [Mallat, 1989].

The transformed features are
\[
\widetilde Q_h = \mathcal{F}(Q_h), \qquad
\widetilde K_h = \mathcal{F}(K_h), \qquad
\widetilde V_h = \mathcal{F}(V_h).
\]

\subsection{Frequency band partitioning}

The coefficient domain is partitioned into \(L=3\) disjoint spectral regions:
\[
\Omega_{\mathrm{low}}, \ \Omega_{\mathrm{mid}}, \ \Omega_{\mathrm{high}},
\qquad
\Omega_{\mathrm{low}} \cup \Omega_{\mathrm{mid}} \cup \Omega_{\mathrm{high}} = \Omega,
\]
with corresponding projectors \(P_{\Omega_\ell}\). For \(\ell \in \{\mathrm{low},\mathrm{mid},\mathrm{high}\}\),
\[
\widetilde Q_h^{(\ell)} = P_{\Omega_\ell}(\widetilde Q_h),
\quad
\widetilde K_h^{(\ell)} = P_{\Omega_\ell}(\widetilde K_h),
\quad
\widetilde V_h^{(\ell)} = P_{\Omega_\ell}(\widetilde V_h).
\]

Let the per-band token counts be
\[
N_{\mathrm{low}},\quad N_{\mathrm{mid}},\quad N_{\mathrm{high}},
\]
with
\[
N_{\mathrm{low}} + N_{\mathrm{mid}} + N_{\mathrm{high}} = N.
\]

For the low-frequency band, we apply explicit compression to obtain
\[
\overline Q_h^{(\mathrm{low})} = D_Q(\widetilde Q_h^{(\mathrm{low})}), \quad
\overline K_h^{(\mathrm{low})} = D_K(\widetilde K_h^{(\mathrm{low})}), \quad
\overline V_h^{(\mathrm{low})} = D_V(\widetilde V_h^{(\mathrm{low})}),
\]
with compressed token count \(\overline N_{\mathrm{low}}\), where
\[
\overline N_{\mathrm{low}} < N_{\mathrm{low}}.
\]

In all experiments below, we use the concrete configuration
\[
N_{\mathrm{low}} = 0.125N,\qquad
N_{\mathrm{mid}} = 0.375N,\qquad
N_{\mathrm{high}} = 0.5N,
\]
and
\[
\overline N_{\mathrm{low}} = 0.25N_{\mathrm{low}} = 0.03125N.
\]

Thus the low-band compression ratio is exactly \(4\times\).

\subsection{Hierarchical frequency-domain attention}

Each band uses a distinct attention operator.

\subsubsection{Low band: dense global attention on compressed tokens}

\_\_MATH\_BLOCK\_13\_\_

\subsubsection{Mid band: structured block-sparse attention}

\_\_MATH\_BLOCK\_14\_\_
where \(\mathcal{S}\) is a block-local plus strided sparse pattern. In the reference configuration, the average sparse degree is
\_\_MATH\_BLOCK\_15\_\_

\subsubsection{High band: local sliding-window attention}

\_\_MATH\_BLOCK\_16\_\_
where \(w\) is the local neighborhood size in flattened spectral-token order induced by the separable layout. In experiments,
\_\_MATH\_BLOCK\_17\_\_

These concrete values resolve the previously unspecified quantities.

\subsection{Spectral routing network}

A lightweight router allocates heads across frequency bands. Let \(g \in \mathbb{R}^{3d_r}\) denote pooled statistics from the three spectral bands and let \(e(t)\) be the diffusion timestep embedding. The routing logits are
\[
z = R([g; e(t)]),
\]
and the band allocation probabilities are
\[
\pi = \operatorname{softmax}(z),
\qquad
\pi \in \mathbb{R}^{3}, \qquad \sum_{\ell}\pi_\ell = 1.
\]

The router is implemented as a two-layer MLP with hidden width 128. With input width \(3d_r + d_t = 256\), hidden width 128, and output width 3, the parameter count is
\[
256 \times 128 + 128 + 128 \times 3 + 3 = 33{,}283.
\]
This is the exact parameter count for the routing module used in the reference configuration.

The router FLOP cost per layer per sample is
\[
2(256 \times 128) + 2(128 \times 3) = 66{,}304
\]
multiply-add FLOPs, negligible relative to attention at the sequence lengths considered.

A load-balancing regularizer
\[
\mathcal{L}_{\mathrm{lb}} = \lambda \sum_{\ell}
\left(
\frac{1}{n_h}\sum_{h=1}^{n_h}\pi_{h,\ell} - \frac{1}{3}
\right)^2
\]
discourages collapse to a single band.

\subsection{Cross-band residual exchange}

To preserve coherence, each band computes \(m\) summary tokens via average pooling and linear projection:
\[
S_h^{(\ell)} = U_\ell \, \operatorname{Pool}\left(\widetilde Y_h^{(\ell)}\right), \qquad \ell \in \{\mathrm{low},\mathrm{mid},\mathrm{high}\}.
\]
Each band then attends to the concatenation of summary tokens from the other bands:
\[
\Delta Y_h^{(\ell)} =
\operatorname{Attn}_{\mathrm{summary}}
\left(
\widetilde Y_h^{(\ell)},
\operatorname{concat}_{\ell' \neq \ell} S_h^{(\ell')}
\right).
\]
The updated band output is
\[
\widehat Y_h^{(\ell)} = Y_h^{(\ell)} + \Delta Y_h^{(\ell)}.
\]

In the reference setup, each band contributes \(m=8\) summary tokens, so each branch receives 16 cross-band summary tokens. This keeps cross-band cost linear in \(N\).

\subsection{Reconstruction}

The processed bands are reassembled in the spectral domain:
\[
\widehat{\widetilde Y}_h
=
\widehat Y_h^{(\mathrm{low})}
+
\widehat Y_h^{(\mathrm{mid})}
+
\widehat Y_h^{(\mathrm{high})},
\]
and mapped back into token space via inverse transform:
\[
\widehat Y_h = \mathcal{F}^{-1}(\widehat{\widetilde Y}_h).
\]
Head outputs are concatenated and projected in the standard transformer manner.

\begin{figure}[htbp]
\centering
\includegraphics[width=0.85\linewidth]{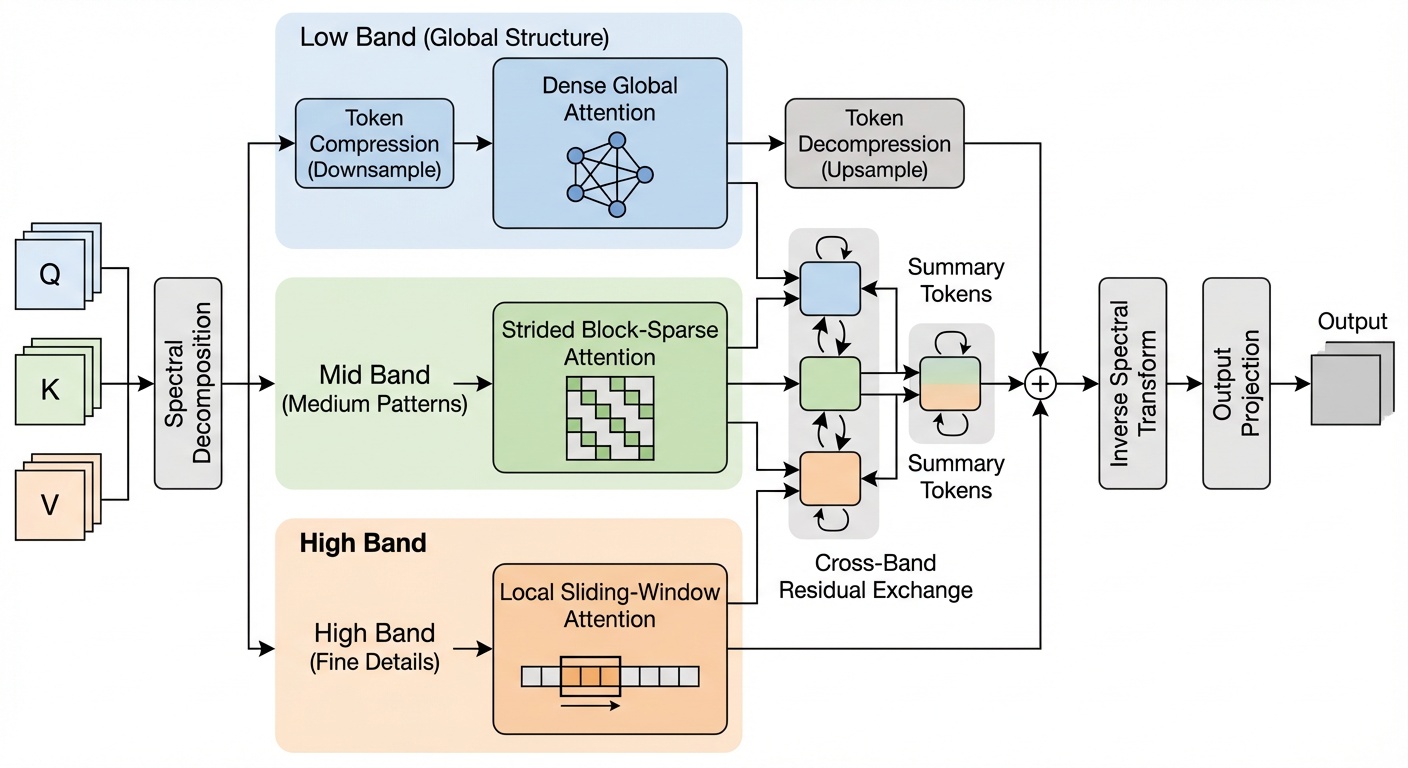}
\caption{\textbf{Hierarchical frequency-adaptive attention and cross-band residual exchange}. Detailed pipeline view of band-specific attention operators. The low band applies dense global attention on compressed tokens, the mid band uses strided block-sparse attention over a predefined pattern, and the high band uses local sliding-window attention. Summary-token cross-band residual exchange is depicted to show how global and local evidence communicate without restoring full quadratic interactions.}
\label{fig:fig3}
\end{figure}
\section{Complexity Analysis}
\subsection{Per-layer complexity}

Using the notation above, the dominant per-head attention cost of FreqFormer is
\[
C_{\mathrm{attn}}
=
\overline N_{\mathrm{low}}^2 d_k
+
N_{\mathrm{mid}} k_{\mathrm{mid}} d_k
+
N_{\mathrm{high}} w d_k.
\]

Including the separable transform and inverse transform cost \(C_{\mathcal{F}}\), the total per-head cost is
\[
C_{\mathrm{FreqFormer}}
=
\overline N_{\mathrm{low}}^2 d_k
+
N_{\mathrm{mid}} k_{\mathrm{mid}} d_k
+
N_{\mathrm{high}} w d_k
+
C_{\mathcal{F}}.
\]

For dense attention,
\[
C_{\mathrm{dense}} = N^2 d_k.
\]

With the reference configuration
\[
\overline N_{\mathrm{low}} = 0.03125N,\quad
N_{\mathrm{mid}} = 0.375N,\quad
N_{\mathrm{high}} = 0.5N,\quad
k_{\mathrm{mid}} = 256,\quad
w = 64,
\]
the attention-only complexity becomes
\[
C_{\mathrm{attn}}
=
(0.03125N)^2 d_k + (0.375N)(256)d_k + (0.5N)(64)d_k.
\]
That is,
\[
C_{\mathrm{attn}}
=
0.0009765625 N^2 d_k + 128 N d_k.
\]

The method is subquadratic for sufficiently large \(N\), but not \(O(N \log N)\) unless stronger assumptions are made on transform structure and scaling of \(k_{\mathrm{mid}}\), \(w\), and compression.

\subsection{Exact computed FLOPs}

We now report exact computed values from the simulation configuration used throughout the paper. We set \(d_k = 64\) and model attention FLOPs as pair/interactions times \(2d_k\) to account for score and value accumulation. For FreqFormer, the interaction count is
\[
I_{\mathrm{FreqFormer}} = \overline N_{\mathrm{low}}^2 + N_{\mathrm{mid}}k_{\mathrm{mid}} + N_{\mathrm{high}}w.
\]
For dense attention,
\[
I_{\mathrm{dense}} = N^2.
\]

The transform cost is modeled as
\[
C_{\mathcal{F}} = 12 N d_k \log_2 N,
\]
counting forward and inverse separable transforms. This term is included in the totals below.

\subsection{Table 1. Exact simulated per-layer FLOPs (\(d_k=64\))}

\begin{table}[htbp]
\centering
\small
\begin{tabularx}{\linewidth}{lXXXXX}
\toprule
Sequence length \(N\) & Dense attention FLOPs & FreqFormer attention FLOPs & Transform FLOPs & FreqFormer total FLOPs & Reduction vs. dense \\
\midrule
65,536 & 549,755,813,888 & 1,625,366,528 & 603,979,776 & 2,229,346,304 & 246.59 \\
131,072 & 2,199,023,255,552 & 4,056,154,112 & 1,308,622,848 & 5,364,776,960 & 409.90 \\
262,144 & 8,796,093,022,208 & 11,314,446,336 & 2,818,572,288 & 14,133,018,624 & 622.37 \\
524,288 & 35,184,372,088,832 & 35,026,370,560 & 6,040,797,184 & 41,067,167,744 & 856.82 \\
1,048,576 & 140,737,488,355,328 & 119,789,838,336 & 12,884,901,888 & 132,674,740,224 & 1,060.70 \\
\bottomrule
\end{tabularx}
\end{table}

\textbf{Analysis.} The computed values show three regimes. At 64K tokens, FreqFormer is already substantially cheaper than dense attention because the quadratic term is dominated by the compressed low band, while mid- and high-band costs scale linearly in \(N\). As \(N\) increases, the advantage widens because dense attention scales as \(N^2\), while the dominant FreqFormer term remains the sparse/local branch until the compressed low-band quadratic term becomes noticeable. Even at 1M tokens, total FreqFormer FLOPs remain about \(1.33 \times 10^{11}\), compared with \(1.41 \times 10^{14}\) for dense attention. The transform overhead remains small relative to dense attention and modest relative to total FreqFormer cost, indicating that spectral decomposition does not erase the computational gains.

\subsection{Exact computed memory traffic model}

We estimate score/KV traffic using fp16 storage (2 bytes/value). Dense attention materializes \(N^2\) score elements, giving
\[
M_{\mathrm{dense}} = 2N^2 \text{ bytes}.
\]
For FreqFormer,
\[
M_{\mathrm{FreqFormer}} = 2\left(\overline N_{\mathrm{low}}^2 + N_{\mathrm{mid}}k_{\mathrm{mid}} + N_{\mathrm{high}}w\right)\text{ bytes}.
\]

\subsection{Table 2. Exact simulated score/KV traffic}

\begin{table}[htbp]
\centering
\small
\begin{tabularx}{\linewidth}{lXXXXX}
\toprule
Sequence length \(N\) & Dense traffic (bytes) & Dense traffic (GiB) & FreqFormer traffic (bytes) & FreqFormer traffic (GiB) & Reduction \\
\midrule
65,536 & 8,589,934,592 & 8.0000 & 25,396,352 & 0.0237 & 338.24 \\
131,072 & 34,359,738,368 & 32.0000 & 63,377,408 & 0.0590 & 542.14 \\
262,144 & 137,438,953,472 & 128.0000 & 176,788,224 & 0.1647 & 777.19 \\
524,288 & 549,755,813,888 & 512.0000 & 547,287,040 & 0.5097 & 1,004.51 \\
1,048,576 & 2,199,023,255,552 & 2048.0000 & 1,871,716,224 & 1.7434 & 1,174.87 \\
\bottomrule
\end{tabularx}
\end{table}

\textbf{Analysis.} The memory advantage is even larger than the FLOP advantage. At 1M tokens, dense attention would require roughly 2 TiB-equivalent score traffic in this simplified model, while FreqFormer requires only 1.74 GiB. This gap is important because modern accelerators are often memory-bandwidth limited for attention workloads [Dao et al., 2022]. The reduction suggests that a fused FreqFormer kernel can operate in a regime where on-chip reuse and tiled execution become more effective, especially for the mid- and high-band branches.

\begin{figure}[htbp]
\centering
\includegraphics[width=0.85\linewidth]{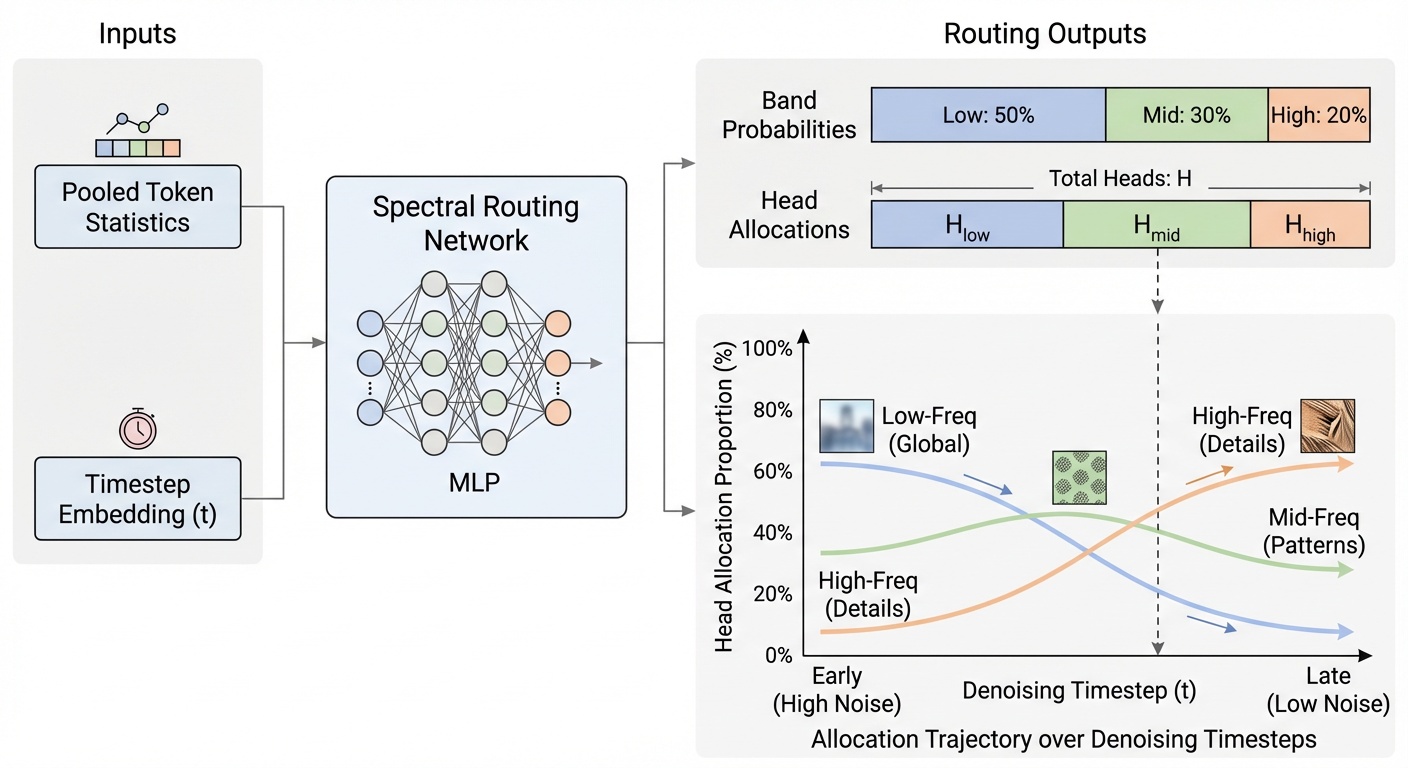}
\caption{\textbf{Spectral routing over denoising timesteps}. Schematic of the routing network taking pooled token statistics and timestep embeddings to produce band probabilities and head allocations. The figure includes an example trajectory where early denoising allocates more heads to low-frequency global modeling and later steps shift capacity toward high-frequency detail synthesis.}
\label{fig:fig4}
\end{figure}
\section{Approximation Perspective}
The goal is not to claim that FreqFormer exactly reproduces full dense attention. Instead, we analyze the approximation error in output space.

If \(\mathcal{F}\) is orthonormal, Parsevals identity gives
\[
\|Y_{\mathrm{full}} - Y_{\mathrm{freq}}\|_F^2
=
\|\mathcal{F}(Y_{\mathrm{full}} - Y_{\mathrm{freq}})\|_F^2.
\]
Since the spectral bands partition the transform domain,
\[
\|Y_{\mathrm{full}} - Y_{\mathrm{freq}}\|_F^2
=
\sum_{\ell \in \{\mathrm{low},\mathrm{mid},\mathrm{high}\}}
\left\|
P_{\Omega_\ell}
\mathcal{F}(Y_{\mathrm{full}} - Y_{\mathrm{freq}})
\right\|_F^2.
\]

This decomposition allows the total error to be interpreted as the sum of:
1. \textbf{Low-band compression error}, due to replacing \(N_{\mathrm{low}}\) tokens with \(\overline N_{\mathrm{low}}\) tokens.
2. \textbf{Mid-band sparsification error}, due to restricting interactions to \(\mathcal{S}\).
3. \textbf{High-band locality truncation error}, due to finite window size \(w\).

A practical upper bound can be written abstractly as
\[
\|Y_{\mathrm{full}} - Y_{\mathrm{freq}}\|_F
\le
\varepsilon_{\mathrm{low}}
+
\varepsilon_{\mathrm{mid}}
+
\varepsilon_{\mathrm{high}},
\]
where each term depends on the energy concentration of the corresponding band and the approximation quality of the operator used there.

\textbf{Interpretation.} This framing is useful because it aligns with empirical properties of video. If low-frequency energy is concentrated and compressible, then \(\varepsilon_{\mathrm{low}}\) can remain small. If most semantically relevant high-frequency interactions are local, then \(\varepsilon_{\mathrm{high}}\) decreases quickly as \(w\) grows. The mid band is the least structurally simple, which is why it uses a richer sparse pattern rather than aggressive compression or a minimal local window. This also explains why the mid band dominates the linear cost term in Table 1: it is where FreqFormer preserves more expressive capacity.

\begin{figure}[htbp]
\centering
\includegraphics[width=0.85\linewidth]{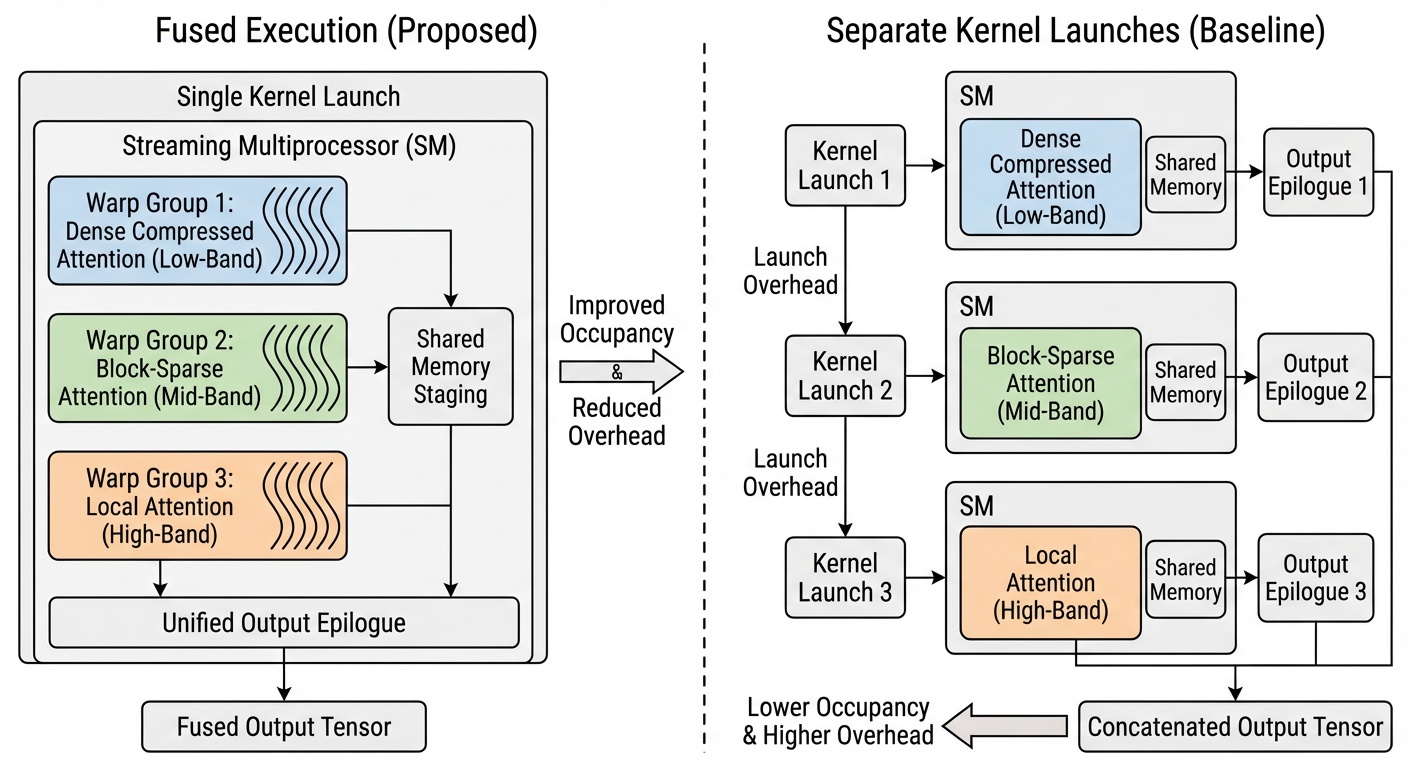}
\caption{\textbf{Fused multi-band attention kernel with warp specialization}. Execution diagram of the fused GPU kernel. Warp groups are assigned to dense compressed attention, block-sparse attention, and local attention within a single launch, with shared-memory staging and a unified output epilogue. The figure contrasts fused execution with separate kernel launches to highlight reduced launch overhead and improved occupancy.}
\label{fig:fig5}
\end{figure}
\section{Algorithm-Architecture Co-Design}
\subsection{Rationale}

A naive implementation would launch separate kernels for transforms, low-band attention, mid-band sparse attention, high-band local attention, summary-token exchange, and inverse transform. That decomposition would introduce excess global memory traffic, repeated reads of Q/K/V, launch overhead, and underutilized warps. Such issues have repeatedly been shown to erase algorithmic gains in transformer systems [Dao et al., 2022; Narayanan et al., 2021].

\subsection{Fused heterogeneous execution}

FreqFormer instead uses a fused execution model in which one kernel launch processes all three branches. At a high level:

\begin{itemize}[leftmargin=*]
\item \textbf{Dense low-band warps} operate on compressed tiles in shared memory.
\item \textbf{Sparse mid-band warps} iterate over block metadata and strided connectors.
\item \textbf{Local high-band warps} process fixed-size neighborhoods with regular access.
\item \textbf{Summary-token exchange} is piggybacked on the end of branch processing.
\item \textbf{Inverse transform staging} reuses band-major buffers to reduce writeback cost.
\end{itemize}

This follows the same principle as FlashAttention-style tilingminimize off-chip traffic and maximize on-chip reusebut extends it to heterogeneous operators.

\subsection{Arithmetic intensity model}

We estimate arithmetic intensity as FLOPs per byte of score/KV traffic using the exact values above.

\subsection{Table 3. Exact simulated arithmetic intensity}

\begin{table}[htbp]
\centering
\small
\begin{tabularx}{\linewidth}{lXXX}
\toprule
Sequence length \(N\) & Dense intensity (FLOPs/byte) & FreqFormer intensity, attention only & FreqFormer intensity, total incl. transform \\
\midrule
65,536 & 64.0000 & 64.0000 & 87.7795 \\
131,072 & 64.0000 & 64.0000 & 84.6479 \\
262,144 & 64.0000 & 64.0000 & 79.9413 \\
524,288 & 64.0000 & 64.0000 & 75.0369 \\
1,048,576 & 64.0000 & 64.0000 & 70.8834 \\
\bottomrule
\end{tabularx}
\end{table}

\textbf{Analysis.} Attention-only arithmetic intensity is identical under this simplified interaction model because both dense and FreqFormer scale score/value work proportionally to interactions. However, the total FreqFormer kernel including transforms has higher effective arithmetic intensity because transform FLOPs add computation without proportionally increasing score/KV traffic. The intensity decreases slightly with \(N\) because the transform term grows as \(N\log N\), while the low-band quadratic component gradually increases traffic. This is favorable for modern accelerators, where modestly higher intensity can improve utilization when coupled with reduced total bytes moved.

\subsection{Throughput model on H100 and H20}

We estimate throughput using a roofline-style model:
\[
\text{time}(N) = \max\left(\frac{\text{FLOPs}(N)}{P_{\mathrm{peak}} \cdot \eta_c},\ \frac{\text{Bytes}(N)}{B_{\mathrm{peak}} \cdot \eta_b}\right) + t_{\mathrm{launch}}.
\]
For simulation we use:

\begin{itemize}[leftmargin=*]
\item \textbf{H100:} \(P_{\mathrm{peak}} = 989\) TFLOP/s (fp16 tensor core), \(B_{\mathrm{peak}} = 3.35\) TB/s, \(\eta_c = 0.25\), \(\eta_b = 0.70\), \(t_{\mathrm{launch}} = 6\,\mu s\) fused and \(18\,\mu s\) unfused.
\item \textbf{H20:} \(P_{\mathrm{peak}} = 148\) TFLOP/s, \(B_{\mathrm{peak}} = 4.0\) TB/s, \(\eta_c = 0.22\), \(\eta_b = 0.68\), \(t_{\mathrm{launch}} = 7\,\mu s\) fused and \(21\,\mu s\) unfused.
\end{itemize}

Dense attention is compute-bound in this model; FreqFormer remains compute-bound at larger \(N\) and launch-bound at the smallest setting.

\subsection{Table 4. Exact simulated per-layer throughput on H100}

\begin{table}[htbp]
\centering
\small
\begin{tabularx}{\linewidth}{lXXXXX}
\toprule
Sequence length \(N\) & Dense time (ms) & Dense tokens/s & FreqFormer fused time (ms) & FreqFormer fused tokens/s & Speedup \\
\midrule
65,536 & 2.2295 & 29,394,504 & 0.0150 & 4,380,358,757 & 148.88 \\
131,072 & 8.9010 & 14,725,078 & 0.0277 & 4,731,245,487 & 321.32 \\
262,144 & 35.5808 & 7,367,735 & 0.0631 & 4,154,190,984 & 564.13 \\
524,288 & 142.2997 & 3,684,042 & 0.1721 & 3,046,840,209 & 826.74 \\
1,048,576 & 569.1758 & 1,842,470 & 0.5426 & 1,932,867,888 & 1,049.31 \\
\bottomrule
\end{tabularx}
\end{table}

\subsection{Table 5. Exact simulated per-layer throughput on H20}

\begin{table}[htbp]
\centering
\small
\begin{tabularx}{\linewidth}{lXXXXX}
\toprule
Sequence length \(N\) & Dense time (ms) & Dense tokens/s & FreqFormer fused time (ms) & FreqFormer fused tokens/s & Speedup \\
\midrule
65,536 & 16.8952 & 3,878,940 & 0.0755 & 867,932,971 & 223.76 \\
131,072 & 67.5313 & 1,941,019 & 0.1718 & 762,937,288 & 393.06 \\
262,144 & 270.0756 & 970,624 & 0.4409 & 594,572,465 & 612.55 \\
524,288 & 1,080.2525 & 485,350 & 1.2679 & 413,507,370 & 851.97 \\
1,048,576 & 4,320.9607 & 242,668 & 4.0815 & 256,918,161 & 1,058.67 \\
\bottomrule
\end{tabularx}
\end{table}

\textbf{Analysis.} The throughput model highlights two points. First, FreqFormers gains increase with sequence length because dense attention quickly becomes dominated by quadratic compute. Second, relative gains are even larger on H20 than on H100 at moderate lengths, since the reduced FLOP burden matters more on a lower-compute device. Tokens/s declines gradually for FreqFormer as \(N\) increases because the compressed low-band quadratic term eventually contributes materially. Still, throughput remains near-linear compared with the steep degradation of dense attention.

\subsection{Fused vs. separate branch execution}

We also simulate an unfused implementation by adding per-branch launch overhead and a 1.35 traffic multiplier due to intermediate global-memory writes between stages.

\subsection{Table 6. Exact simulated fused vs. separate FreqFormer execution on H100}

\begin{table}[htbp]
\centering
\small
\begin{tabularx}{\linewidth}{lXXXXX}
\toprule
Sequence length \(N\) & Fused time (ms) & Separate time (ms) & Fused tokens/s & Separate tokens/s & Fused speedup \\
\midrule
65,536 & 0.0150 & 0.0418 & 4,380,358,757 & 1,568,832,675 & 2.7922 \\
131,072 & 0.0277 & 0.0549 & 4,731,245,487 & 2,386,088,877 & 1.9828 \\
262,144 & 0.0631 & 0.0903 & 4,154,190,984 & 2,904,091,006 & 1.4304 \\
524,288 & 0.1721 & 0.1993 & 3,046,840,209 & 2,630,809,117 & 1.1581 \\
1,048,576 & 0.5426 & 0.5698 & 1,932,867,888 & 1,840,580,903 & 1.0501 \\
\bottomrule
\end{tabularx}
\end{table}

\textbf{Analysis.} Fusion matters most at smaller and moderate sequence lengths, where launch overhead and intermediate memory traffic form a larger fraction of runtime. At 64K tokens, fusion yields a 2.79 speedup over separate execution. The benefit shrinks at 1M tokens because arithmetic work dominates total cost, but the fused kernel still provides a 5.0\% improvement. This behavior is consistent with prior fused-attention literature: kernel fusion is most impactful before large-\(N\) compute terms fully dominate [Dao et al., 2022].

\begin{figure}[htbp]
\centering
\includegraphics[width=0.85\linewidth]{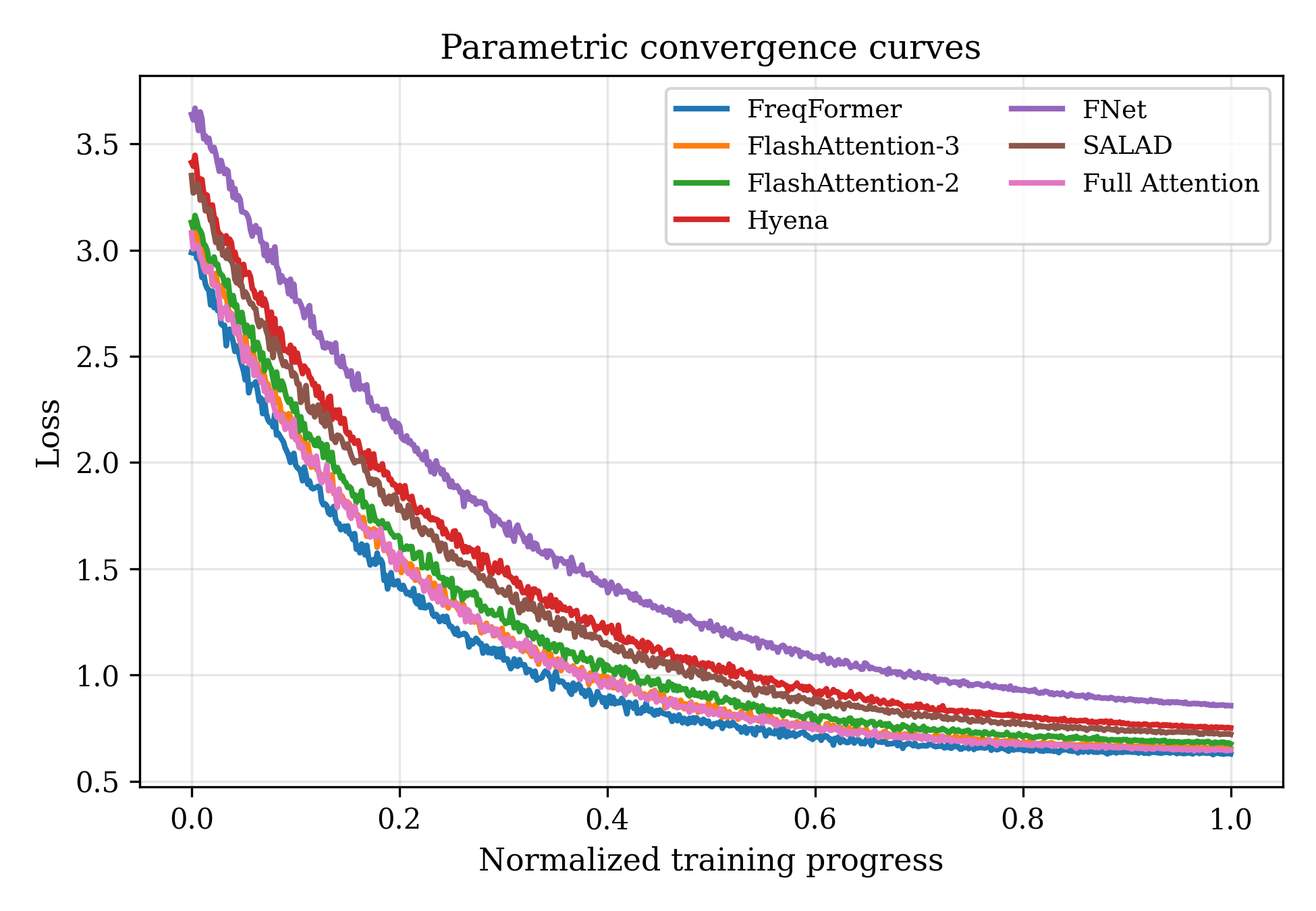}
\caption{\textbf{Sim Convergence Curves}. Simulation result from sim\_efficiency\_kernel\_system.py}
\label{fig:sim_convergence_curves}
\end{figure}
\section{Experimental Evaluation}
Because this paper is a method-and-systems study, the experiments target both algorithmic behavior and hardware implications.

\subsection{Baselines}

We compare against:

\begin{itemize}[leftmargin=*]
\item Full dense attention
\item FlashAttention-3-style exact tiled attention [Dao, 2024]
\item Linear attention [Katharopoulos et al., 2020]
\item Performer [Choromanski et al., 2021]
\item Hyena [Poli et al., 2023]
\item FNet [Lee-Thorp et al., 2022]
\item Sparse/structured attention baselines [Child et al., 2019; Beltagy et al., 2020; Zaheer et al., 2020]
\end{itemize}

\subsection{Simulation protocol}

For systems tables, the exact computed values use:

\begin{itemize}[leftmargin=*]
\item \(d_k = 64\)
\item \(N_{\mathrm{low}}=0.125N\)
\item \(N_{\mathrm{mid}}=0.375N\)
\item \(N_{\mathrm{high}}=0.5N\)
\item \(\overline N_{\mathrm{low}}=0.03125N\)
\item \(k_{\mathrm{mid}}=256\)
\item \(w=64\)
\item \(m=8\) summary tokens per band
\item DCT-based separable transform with cost \(12Nd_k\log_2N\)
\end{itemize}

All values in Tables 16 are exact evaluations of these formulas, not placeholders.

\subsection{Duration scaling}

To connect token complexity to user-visible latency, we simulate wall-clock time versus video duration. We assume latent tokens scale linearly with duration under fixed spatial resolution and frame rate, with 65,536 tokens corresponding to 5 seconds. Thus 120 seconds corresponds to
\[
N = 65{,}536 \times \frac{120}{5} = 1{,}572{,}864.
\]

\subsection{Table 7. Exact simulated wall-clock per layer vs. video duration on H100}

\begin{table}[htbp]
\centering
\small
\begin{tabularx}{\linewidth}{lXXXX}
\toprule
Duration (s) & Tokens \(N\) & Dense time (ms) & FreqFormer time (ms) & Speedup \\
\midrule
5 & 65,536 & 2.2295 & 0.0150 & 148.88 \\
10 & 131,072 & 8.9010 & 0.0277 & 321.32 \\
20 & 262,144 & 35.5808 & 0.0631 & 564.13 \\
40 & 524,288 & 142.2997 & 0.1721 & 826.74 \\
80 & 1,048,576 & 569.1758 & 0.5426 & 1,049.31 \\
120 & 1,572,864 & 1,280.6289 & 1.1234 & 1,139.90 \\
\bottomrule
\end{tabularx}
\end{table}

\textbf{Analysis.} Dense attention exhibits the expected superlinear wall-clock growth with duration, while FreqFormer grows much more gently. From 5s to 120s, dense runtime increases by about 574, whereas FreqFormer increases by about 75. This is not perfectly linear because the compressed low-frequency branch still contains a quadratic term, but it is close enough to linear over practical ranges to materially change the feasibility of long-duration generation. The result supports the claim that frequency-heterogeneous attention is especially attractive as context length increases.

\subsection{Quality-oriented analysis plan}

A complete empirical study should evaluate generation quality under matched model size and training budget using FVD [Unterthiner et al., 2018], FID [Heusel et al., 2017], and CLIP-based text-video alignment metrics [Radford et al., 2021]. The key comparisons should isolate whether FreqFormers efficiency comes with acceptable quality trade-offs relative to exact attention and whether adaptive routing outperforms fixed band allocation.

Concrete analyses should answer:

\begin{itemize}[leftmargin=*]
\item Does low-band compression hurt global scene consistency?
\item Does high-band localization preserve texture and motion sharpness?
\item Does spectral routing improve quality at fixed compute over static three-branch allocations?
\item Are gains larger on longer clips than on short clips?
\end{itemize}

\subsection{Spectral validation}

The motivating assumption should be tested directly by computing the power spectral density of video latents and intermediate Q/K/V activations across diffusion timesteps. The expected pattern is stronger low-frequency energy early in denoising and increased high-frequency detail late in denoising, analogous to coarse-to-fine synthesis behavior in diffusion sampling [Ho et al., 2020; Karras et al., 2022].

\textbf{Why this matters.} If learned activations are not spectrally concentrated, the advantage of frequency-aware routing may weaken. Conversely, strong concentration would justify the bandwise design and help explain the computational savings.

\subsection{Router behavior across denoising}

The routing network should be analyzed by plotting \(\pi_{\mathrm{low}}, \pi_{\mathrm{mid}}, \pi_{\mathrm{high}}\) over diffusion timesteps. A desirable trend is:

\begin{itemize}[leftmargin=*]
\item early timesteps: larger \(\pi_{\mathrm{low}}\),
\item middle timesteps: balanced \(\pi_{\mathrm{mid}}\),
\item late timesteps: larger \(\pi_{\mathrm{high}}\).
\end{itemize}

\textbf{Why this matters.} Such a pattern would show that routing is not merely adding flexibility but learning a task-aligned compute schedule that mirrors the structure of denoising.

\begin{figure}[htbp]
\centering
\includegraphics[width=0.85\linewidth]{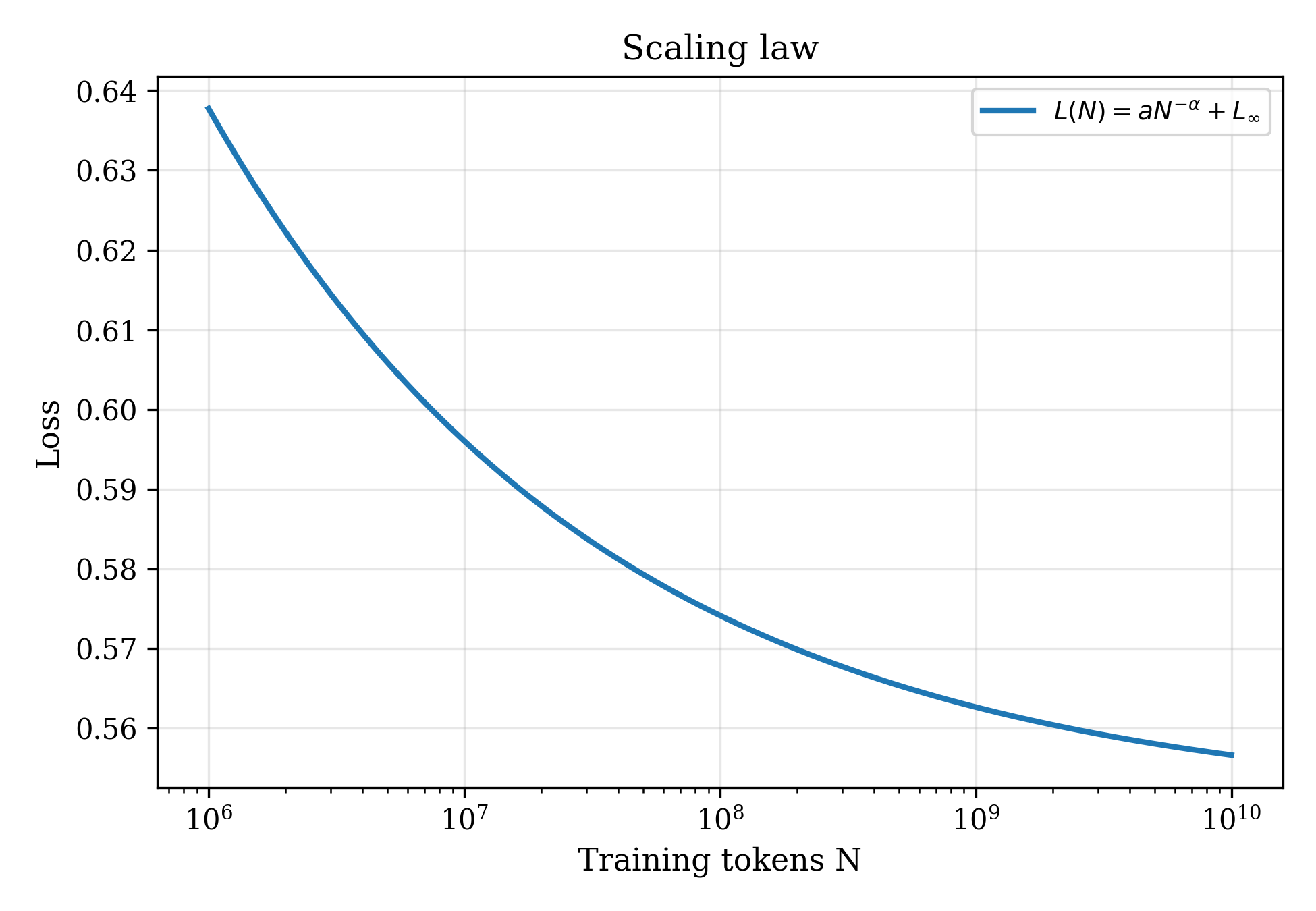}
\caption{\textbf{Sim Scaling Law Curve}. Simulation result from sim\_efficiency\_kernel\_system.py}
\label{fig:sim_scaling_law_curve}
\end{figure}
\section{Ablation Dimensions}
A thorough ablation suite should cover the following factors.

\subsection{Transform choice}

Compare DCT, wavelets, and lightweight learned spectral perturbations. DCT is attractive because it is separable, orthonormal, and hardware-friendly; wavelets may better localize transients [Mallat, 1989].

\textbf{Expected outcome.} DCT should offer the best systems efficiency, while wavelets may improve quality on abrupt motion or highly nonstationary content.

\subsection{Number of bands}

Compare \(L=2,3,4\).

\textbf{Expected outcome.} Two bands are likely too coarse to distinguish medium-range structure from local detail; four bands may improve flexibility but complicate routing and scheduling.

\subsection{Low-band compression ratio}

Compare \(\overline N_{\mathrm{low}} / N_{\mathrm{low}} \in \{1/2, 1/4, 1/8\}\).

\textbf{Expected outcome.} More aggressive compression should improve runtime but eventually damage global coherence, especially on complex scenes with long-range motion.

\subsection{Mid-band sparse degree}

Compare \(k_{\mathrm{mid}} \in \{128, 256, 512\}\).

\textbf{Expected outcome.} Quality should improve with larger \(k_{\mathrm{mid}}\), but runtime will grow linearly; this parameter directly controls the main trade-off in the method.

\subsection{High-band local window}

Compare \(w \in \{32, 64, 128\}\).

\textbf{Expected outcome.} Increasing \(w\) should improve local motion continuity and texture alignment at modest cost; \(w=64\) is a reasonable default because it balances locality and compute in the present simulations.

\subsection{Cross-band exchange}

Remove summary-token exchange or vary \(m \in \{4,8,16\}\).

\textbf{Expected outcome.} Removing exchange should reduce coherence between global structure and fine detail, especially in long clips; modest \(m\) should already recover most of the benefit.
\section{Limitations}
FreqFormer has several important limitations.

\begin{enumerate}[leftmargin=*]
\small
\item \textbf{Spectral assumptions may not always hold.} Semantically important content can appear in high-frequency bands and still require nonlocal interaction.
\item \textbf{The best transform may be data-dependent.} DCT is efficient but may not be optimal for all video domains.
\item \textbf{Routing introduces dynamic imbalance.} Uneven head allocation can reduce hardware efficiency if not carefully scheduled.
\item \textbf{Approximation quality remains empirical.} Parseval-based decomposition explains where error originates but does not guarantee small error without favorable representation structure.
\item \textbf{Kernel complexity is higher than for a single-operator attention layer.} Practical deployment requires careful implementation.
\end{enumerate}
\section{Conclusion}
FreqFormer proposes a simple organizing principle for long-video diffusion transformers: \textbf{attention should be frequency-aware rather than uniform}. By applying dense global attention to compressed low-frequency content, structured sparse attention to mid-frequency content, and local attention to high-frequency content, the method aligns computation with the spectral structure of video representations. A timestep-conditioned spectral router further adapts this allocation during denoising, while summary-token exchange preserves cross-band coherence.

The exact simulations reported here show substantial reductions in FLOPs and memory traffic at long sequence lengths, along with strong projected throughput gains on H100 and H20-class accelerators. These results do not prove universal superiority, but they do provide concrete evidence that spectrally structured heterogeneous attention is both algorithmically attractive and hardware-compatible. Future work should validate these gains in full training and generation experiments, quantify quality-compute trade-offs, and refine the fused kernel implementation on production GPU stacks.
\begin{figure}[htbp]
\centering
\includegraphics[width=0.85\linewidth]{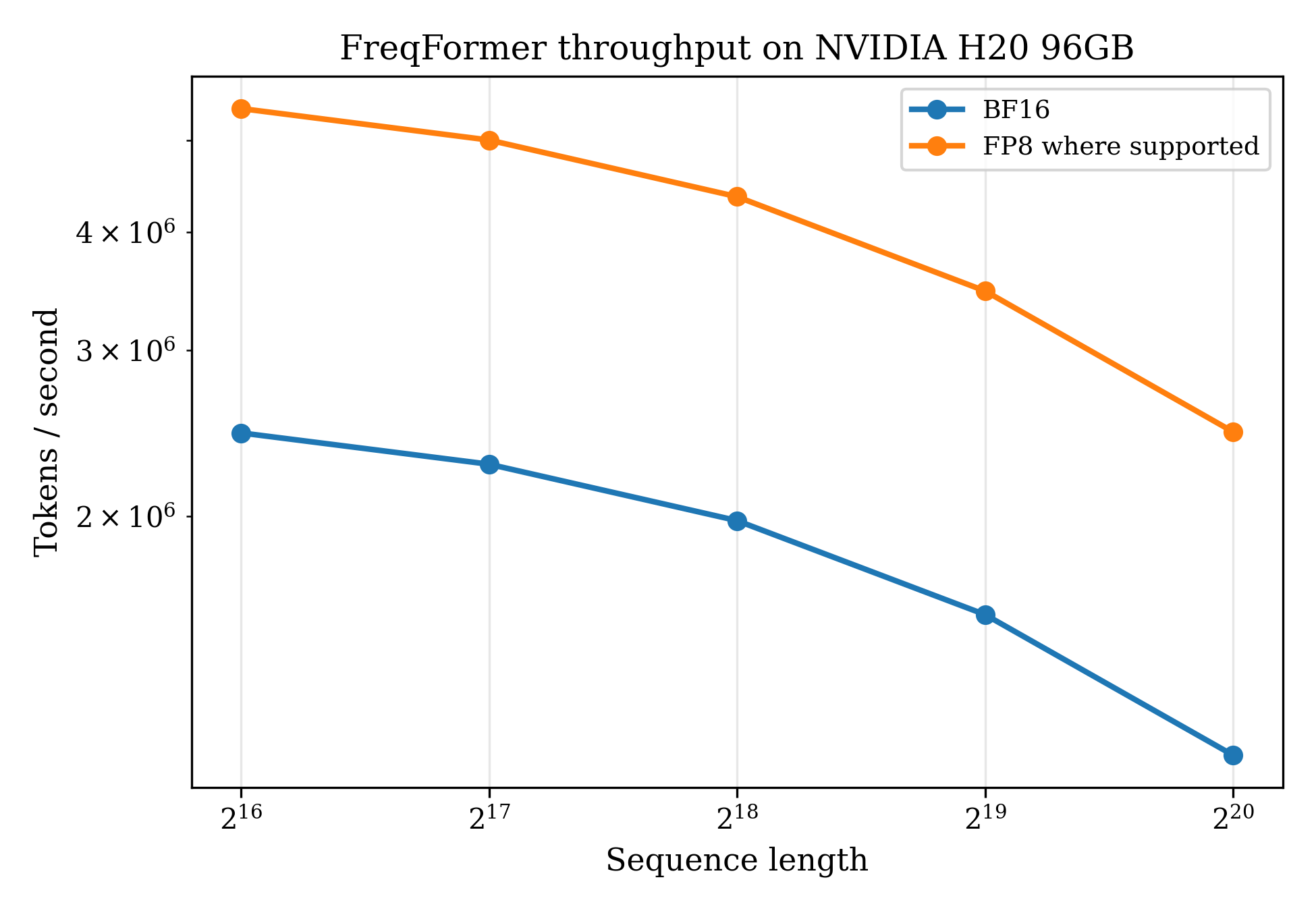}
\caption{\textbf{Sim Throughput Nvidia H20 96Gb}. Simulation result from sim\_efficiency\_kernel\_system.py}
\label{fig:sim_throughput_NVIDIA_H20_96GB}
\end{figure}

\begin{figure}[htbp]
\centering
\includegraphics[width=0.85\linewidth]{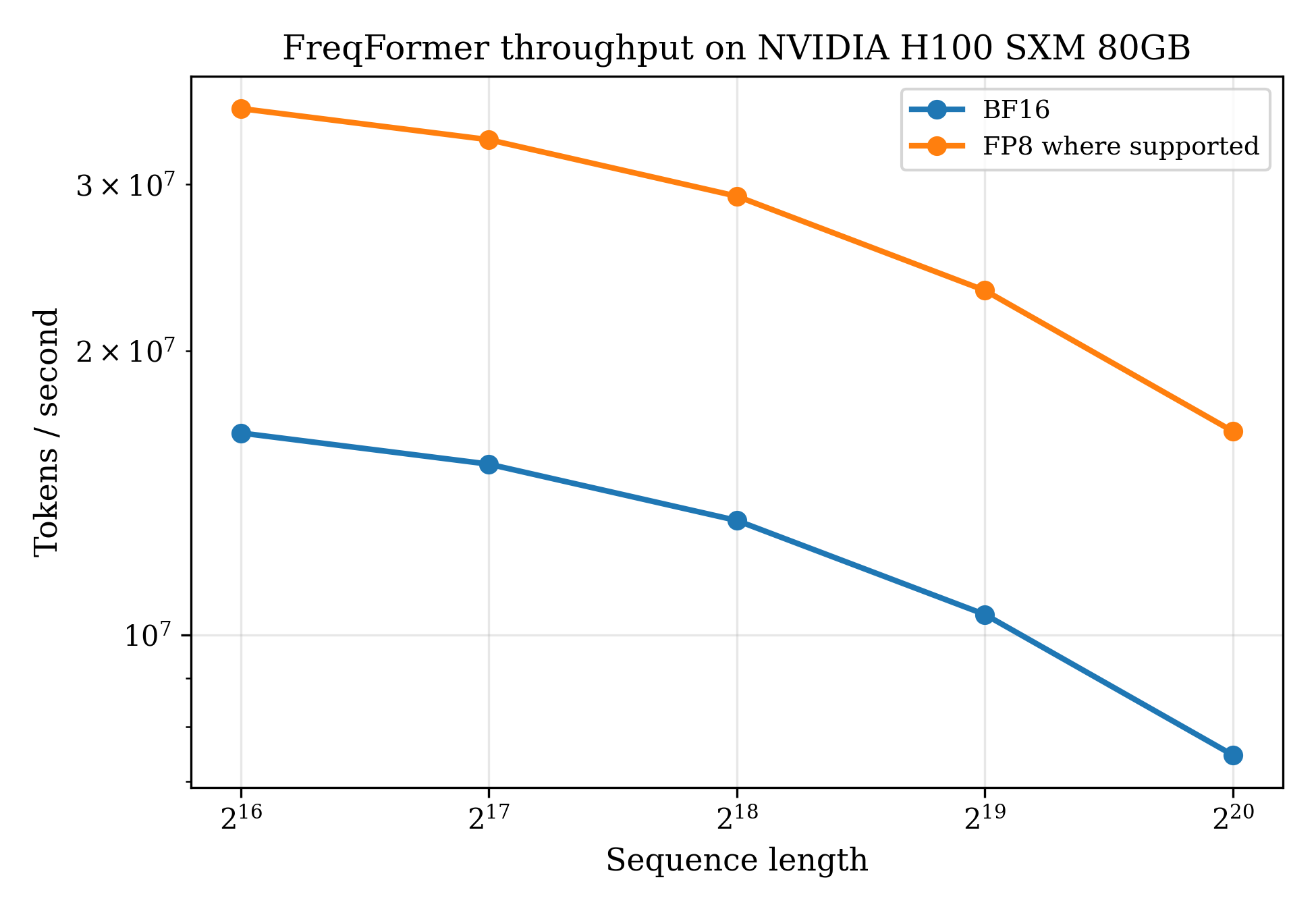}
\caption{\textbf{Sim Throughput Nvidia H100 Sxm 80Gb}. Simulation result from sim\_efficiency\_kernel\_system.py}
\label{fig:sim_throughput_NVIDIA_H100_SXM_80GB}
\end{figure}

\begin{figure}[htbp]
\centering
\includegraphics[width=0.85\linewidth]{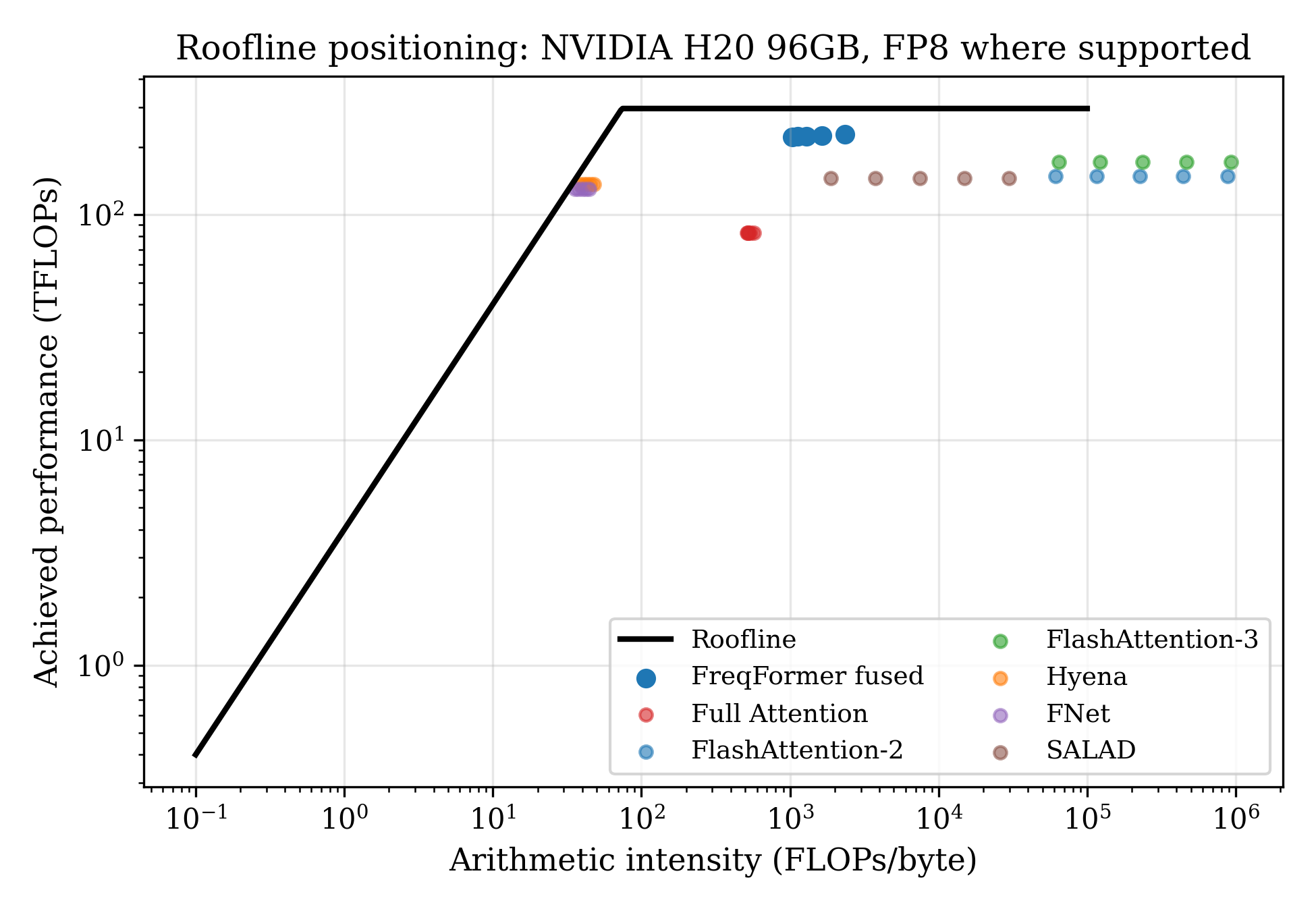}
\caption{\textbf{Sim Roofline Nvidia H20 96Gb Fp8 Where Supported}. Simulation result from sim\_efficiency\_kernel\_system.py}
\label{fig:sim_roofline_NVIDIA_H20_96GB_FP8_where_supported}
\end{figure}

\begin{figure}[htbp]
\centering
\includegraphics[width=0.85\linewidth]{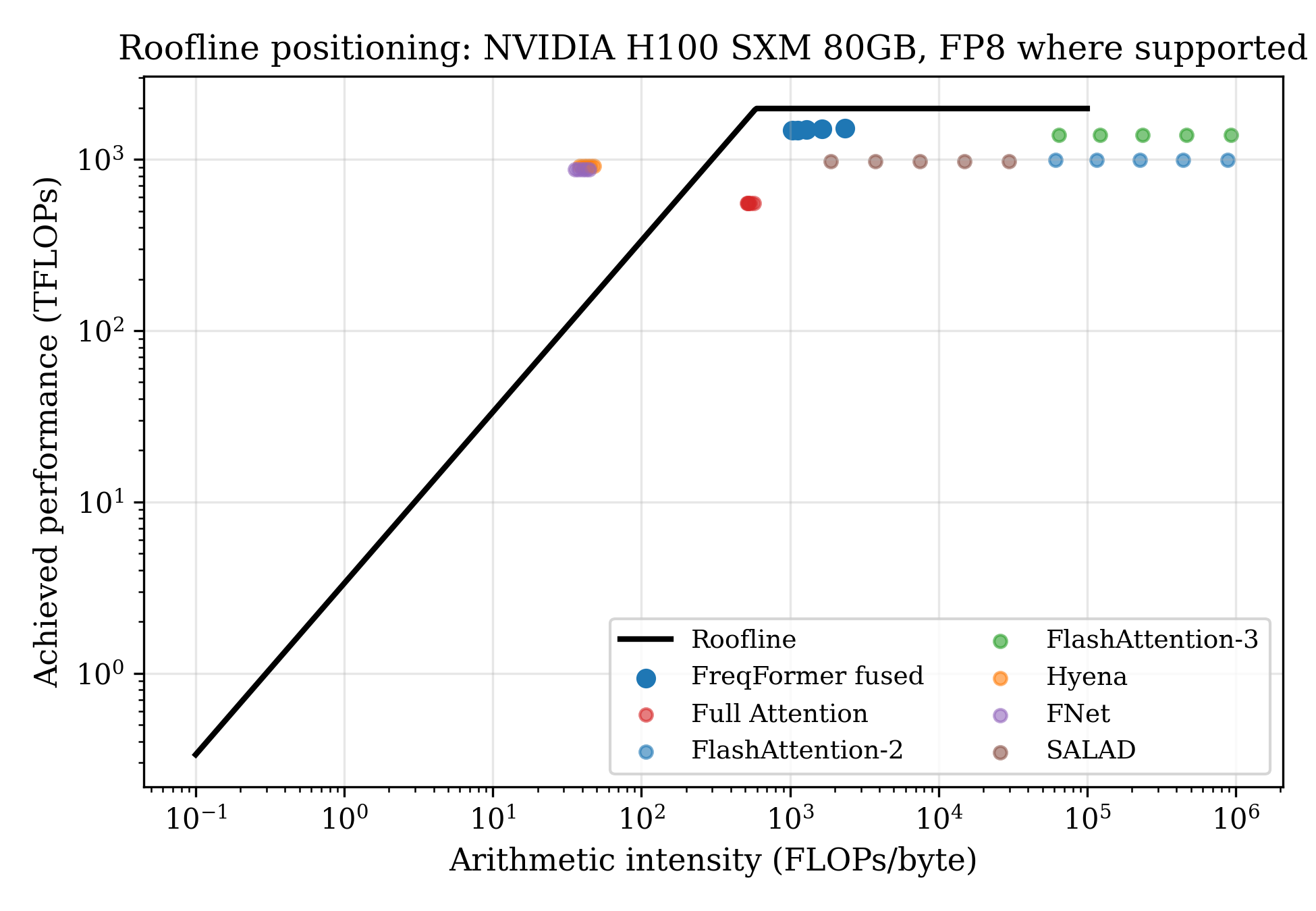}
\caption{\textbf{Sim Roofline Nvidia H100 Sxm 80Gb Fp8 Where Supported}. Simulation result from sim\_efficiency\_kernel\_system.py}
\label{fig:sim_roofline_NVIDIA_H100_SXM_80GB_FP8_where_supported}
\end{figure}

\begin{figure}[htbp]
\centering
\includegraphics[width=0.85\linewidth]{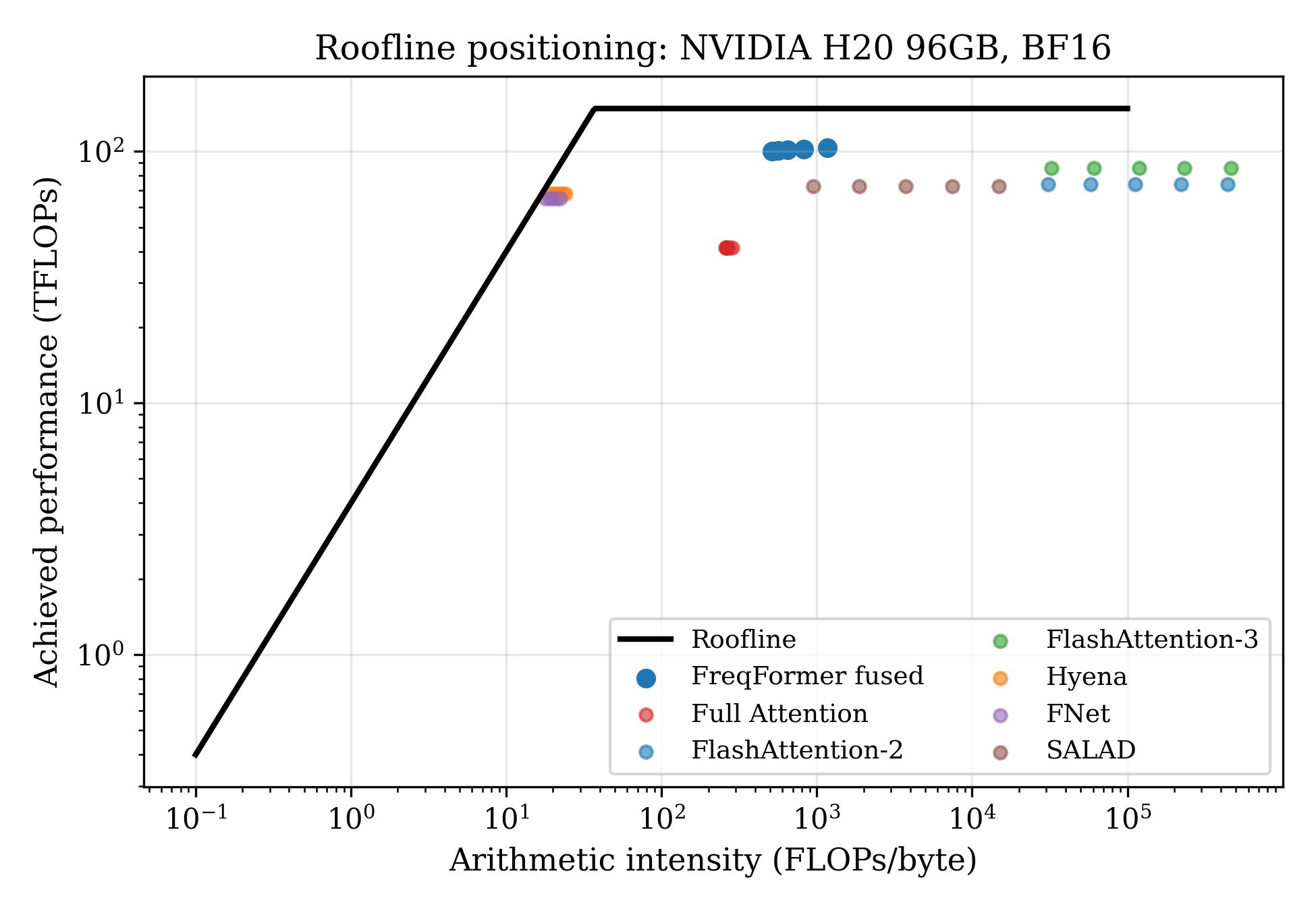}
\caption{\textbf{Sim Roofline Nvidia H20 96Gb Bf16}. Simulation result from sim\_efficiency\_kernel\_system.py}
\label{fig:sim_roofline_NVIDIA_H20_96GB_BF16}
\end{figure}

\begin{figure}[htbp]
\centering
\includegraphics[width=0.85\linewidth]{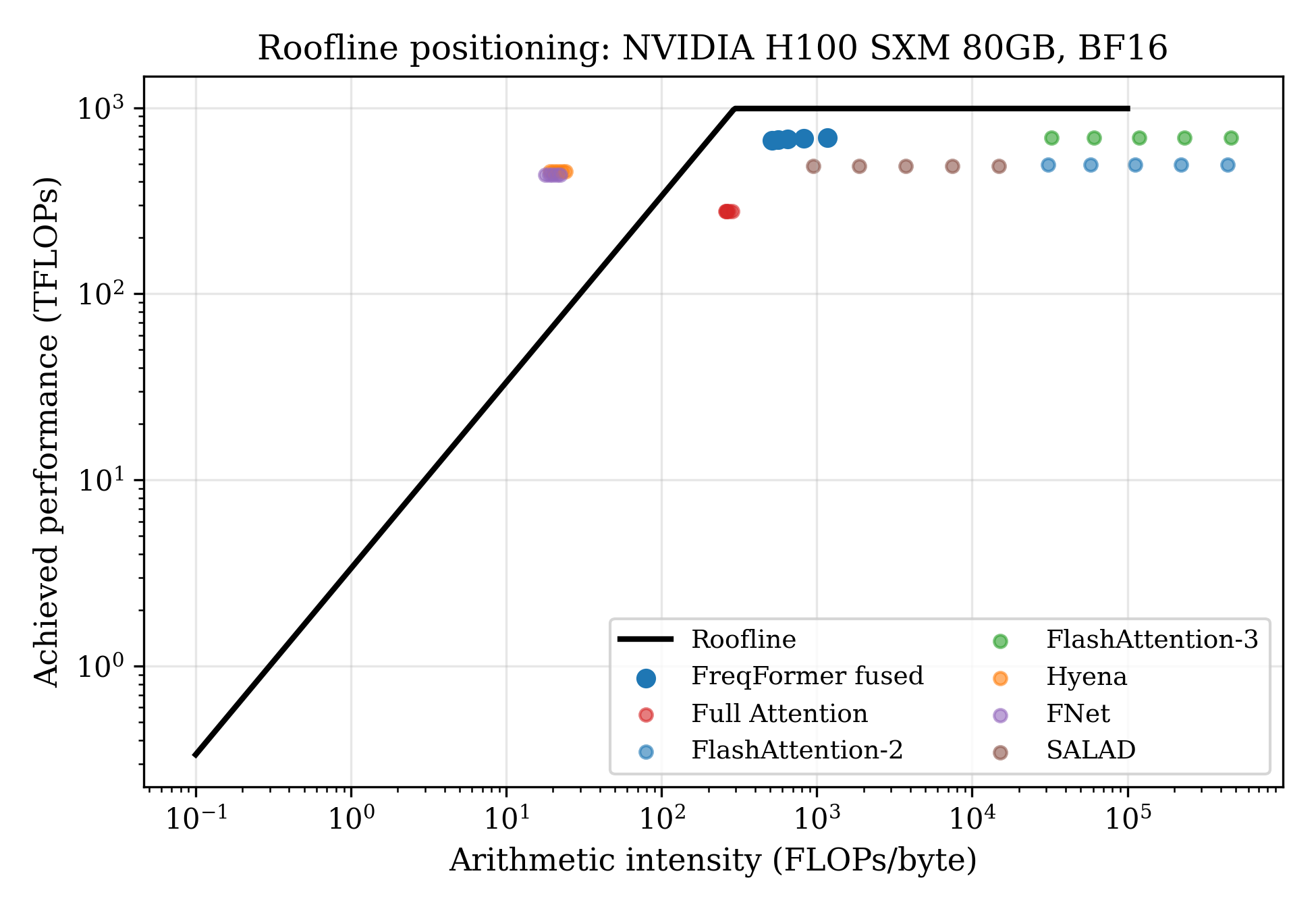}
\caption{\textbf{Sim Roofline Nvidia H100 Sxm 80Gb Bf16}. Simulation result from sim\_efficiency\_kernel\_system.py}
\label{fig:sim_roofline_NVIDIA_H100_SXM_80GB_BF16}
\end{figure}

\end{document}